\documentclass{article}
\usepackage{spconf}
\usepackage{amsfonts}
\usepackage{amssymb}
\usepackage{bm}
\usepackage{enumitem}
\usepackage{float}
\usepackage{url}
\usepackage{multirow}
\usepackage{subcaption}
\usepackage{cite}
\usepackage{amsmath,amssymb,amsfonts}
\usepackage{graphicx}
\usepackage{textcomp}
\usepackage{xcolor}
\usepackage{booktabs}
\usepackage{float}
\usepackage{adjustbox}
\usepackage{hyperref}
\usepackage{comment}
\usepackage{algorithm}
\usepackage{algpseudocode} 
\usepackage[table]{xcolor}
\usepackage[font=footnotesize,skip=2pt]{caption}
\usepackage{bbm}
\usepackage{arydshln}

\makeatletter
\newcommand{\StatePar}[1]{%
  \State\parbox[t]{\dimexpr\linewidth-\ALG@thistlm}{\strut #1\strut}%
}
\makeatother

\makeatletter
\renewcommand{\paragraph}{\@startsection{paragraph}{4}{\z@}%
  {0.75ex \@plus 0.5ex \@minus 0.2ex}
  {-1em}
  {\normalfont\normalsize\bfseries}}
\makeatother
\makeatletter
\renewcommand{\ALG@beginalgorithmic}{\small} 
\makeatother

\algrenewcommand\algorithmicrequire{\textbf{Input:}}
\algrenewcommand\algorithmicensure{\textbf{Output:}}
\algrenewcommand\algorithmiccomment[1]{\hfill\(\triangleright\) #1}
\makeatletter
\g@addto@macro\normalsize{%
  \setlength\abovedisplayskip{3pt}%
  \setlength\belowdisplayskip{3pt}%
  \setlength\abovedisplayshortskip{1pt}%
  \setlength\belowdisplayshortskip{1pt}%
}
\makeatother

\makeatletter
\renewcommand{\section}{\@startsection{section}{1}{\z@}%
  {1.5ex}
  {1ex}
  {\normalfont\normalsize\bfseries}} 
\makeatother


\title{WiSPER: Pose-Supervised Predictive and Residual Flow Refinement for Multi-Person 3D Pose Estimation with WiFi CSI}
\name{
\begin{tabular}{c}
Gabriel Jun Rong Lee$^{1}$ \qquad Shanhong Liu$^{2}$ \qquad   Pai Chet Ng$^{1}$   
\qquad Konstantinos N. Plataniotis$^{3}$ \\
\qquad  Jamal Seyedmohammadi$^{1}$ \qquad S. Mohammad Sheikholeslami$^{3}$ 
\end{tabular}
}
\address{
$^{1}$Infocomm Technology Cluster, Singapore Institute of Technology, Singapore \\
$^{2}$Information Systems Technology and Design, Singapore University of Technology and Design \\
$^{3}$Department of Electrical and Computer Engineering, University of Toronto, Canada \\
}

\begin{document}
\ninept

\maketitle

\begin{abstract} 
Multi-person 3D pose estimation with WiFi channel state
information (CSI) is challenging because reflections from
different people overlap without directly identifying
individual joints. Existing masked embedding objectives
capture wireless relationships without explicit pose
supervision, while structured decoders can retain
coordinate errors. We propose WiSPER, a two-stage
framework combining pose-aware predictive pretraining
with conditional residual flow refinement.
Pose-Aware Masked Embedding Learning (PAMEL) couples
masked latent prediction with auxiliary pose-set
supervision on the same CSI context, guiding
the encoder toward joint localization from partial
observations. Residual Flow refinement with Transformer
(ReFT) generates a set of pose candidates to accommodate
a variable number of people and refines each candidate
through a conditional flow guided by its coarse
coordinates and per-joint decoder features.
Both stages use paired CSI and pose annotations during
training, while inference requires only CSI.
Experiments on the PiW3D dataset show that WiSPER achieves
an overall mean per-joint position error of 63.72~mm,
a 40.0\% reduction relative to WiFi-JEPA.
For experiments with two- and three-person,
WiSPER reduces MPJPE by 42.1\% and 38.1\%, respectively.
Pose-supervised pretraining configurations obtain
lower errors than CSI-only JEPA, and enabling the
trained residual refiner reduces overall MPJPE by
13.8--15.6\% across the evaluated configurations.
\end{abstract}


\section{Introduction} \label{sec:intro}
Human pose estimation~\cite{liu2022recent,zheng2023deep}
provides a structured description of body configuration
for rehabilitation, exercise monitoring, and
human--computer interaction. Recent studies investigate
spatio-temporal expert mixtures~\cite{11464440}
and conditional random fields~\cite{11460646}
for 3D pose estimation. Other developments address
cross-dataset adaptation through skeleton prompt
tuning~\cite{10890436}, camera perspective in
monocular 3D estimation~\cite{Hao_2025_ICCV}, and
efficient image-based inference through dynamic
model routing~\cite{Xu_2025_CVPR}. These advances
improve different aspects of pose estimation, but
visual monitoring remains constrained by lighting,
occlusion, and concerns about recording identifiable
data. WiFi sensing offers an alternative when visual
observations are unavailable or undesirable.
It relies on channel state information
(CSI)~\cite{ratnam2024optimal,11498405}, which describes the
complex-valued channel response across frequency subcarriers
and antenna links. Human bodies reflect and obstruct wireless
signals, changing the amplitude and phase of CSI and providing
information about body position and movement across links
and time~\cite{ahmad2024wifi,miao2025wi}. Recovering multiple
poses is particularly challenging because reflections from
different people and the environment are combined within
the same measurements. 
Yet, the signal variations correspond to a particular person or joint,
and different links provide unequal information about each
body. 

Recent work has advanced both wireless representations and
pose decoding. GoPose~\cite{gopose3534605} uses
two-dimensional angle-of-arrival spectra with spatial and
temporal learning to recover 3D skeletons, while
Person-in-WiFi 3D~\cite{person3dyan} introduces
Transformer-based set prediction for multi-person estimation.
DT-Pose~\cite{chen2026dt} combines self-supervised
representation learning with topology-constrained decoding,
and GraphPose-Fi~\cite{chen2026graph} models joint
relationships through antenna attention and graph-based
regression. More recently, WiFi-JEPA~\cite{kim2026wifijepa}
learns cross-link representations through masked latent
prediction before supervised pose estimation. Its pretraining
objective learns from CSI without pose annotations.
When paired annotations are available, a complementary
approach is to introduce pose supervision during predictive
pretraining, requiring the visible-context representation
to support both latent prediction and joint localization.
At the decoding stage, hierarchical pose
Transformers~\cite{shi2022end} produce structured pose
estimates whose remaining coordinate errors can be modeled
explicitly. This motivates studying pose-supervised
predictive initialization together with residual refinement
for multi-person CSI-based pose estimation.

We propose WiSPER, a two-stage framework that combines
these two design choices. Stage~1, Pose-Aware Masked
Embedding Learning (PAMEL), is supervised predictive
pretraining: it couples pose-set supervision with a
JEPA-style masked latent objective on the same visible
CSI context. Pose annotations guide joint localization,
while the latent objective encourages prediction across
wireless links. Stage~2, Residual Flow refinement with
Transformer (ReFT), transfers the encoder to a hierarchical
pose Transformer and learns conditional corrections to
its pose candidates. Each correction depends on the
candidate's coarse coordinates and per-joint decoder
features, connecting the current body estimate with
CSI-dependent information. Both stages use the same
paired training split; the first stage provides additional
supervised optimization before downstream pose estimation.
Our contributions are:
\begin{itemize}
    \item We formulate PAMEL as supervised predictive
    pretraining, combining pose-set supervision and
    masked embedding prediction on the same partial
    CSI observations.

    \item We develop ReFT, which conditions an
    endpoint-parameterized residual flow on coarse
    pose coordinates and per-joint Transformer features
    to refine pose candidates. 

    \item Our evaluations on PiW3D dataset show that applying the
    trained residual refiner reduces overall MPJPE by
    13.8--15.6\% across the three evaluated pretraining
    configurations. 
\end{itemize}

\section{Proposed Method} \label{sec:method}

\begin{figure*}[t]
 \centering
 \includegraphics[width=\textwidth]{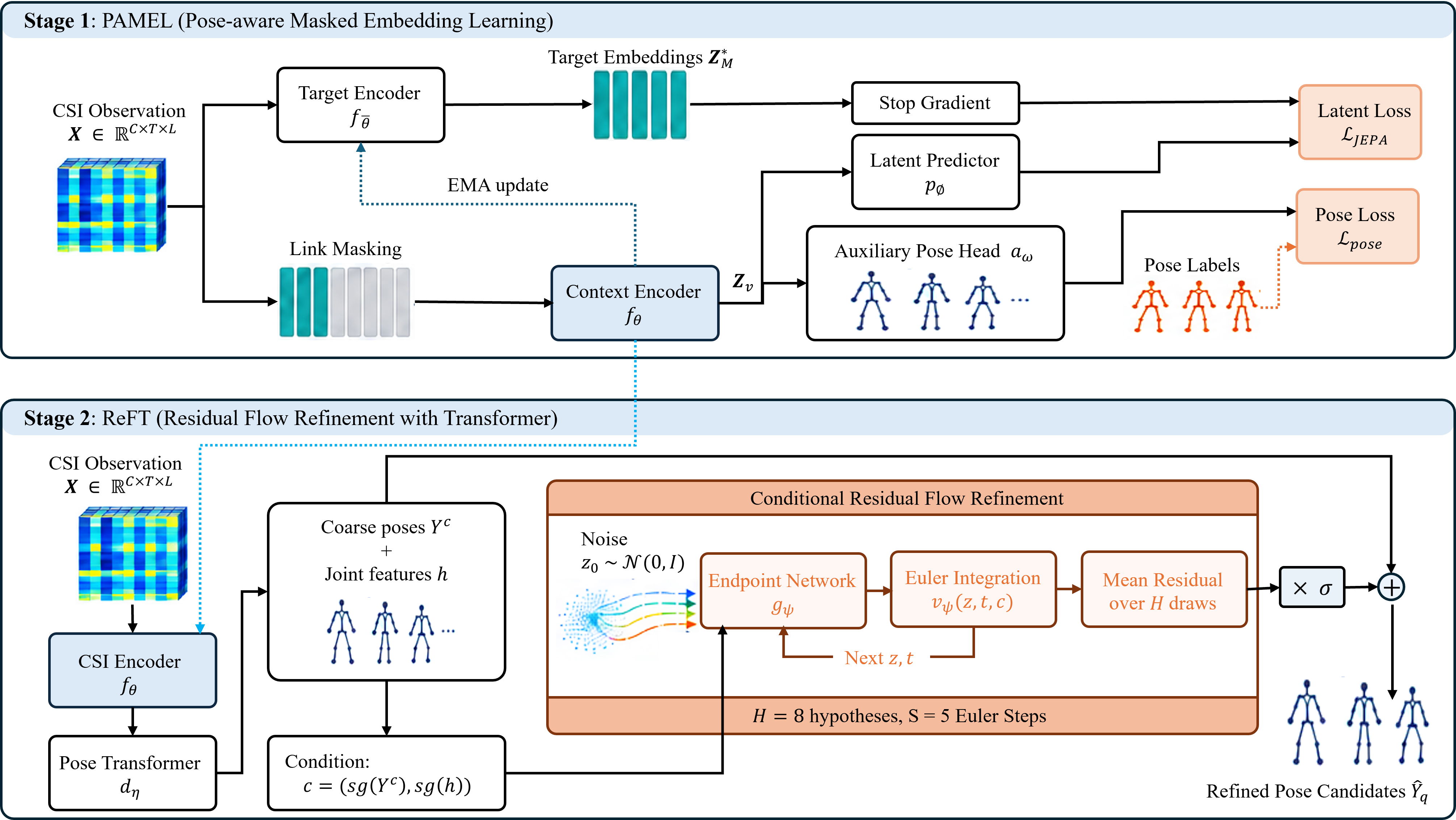}
 \caption{Our WiSPER consists of 2 stages: Stage~1, PAMEL, combines masked
latent prediction with auxiliary pose supervision on visible
CSI context. Stage~2, ReFT, uses the pretrained encoder and
a pose Transformer to generate coarse pose candidates,
then refines them through conditional residual flow guided
by their coordinates and joint features. Residuals from
$H$ sampled trajectories are averaged, scaled by $\sigma$,
and added to each coarse pose.}
 \label{fig:overview}
\end{figure*}

\subsection{Problem formulation}
We consider multi-person 3D pose estimation from WiFi CSI.
Let $\mathbf{X}\in\mathbb{R}^{C\times T\times L}$ denote a
CSI observation, where $C$, $T$, and $L$ are the numbers of
signal channels, temporal samples, and wireless links, respectively.
Each observation is associated with $P$ people.
The annotated pose of person $p$ is 
$\mathbf{Y}_p\in\mathbb{R}^{K\times 3}$, whose rows contain
the 3D coordinates of $K$ predefined body joints, such as
the shoulders, elbows, and wrists.
The annotated poses form the set
\begin{equation}
    Y=\left\{\mathbf{Y}_p\right\}_{p=1}^{P}.
    \label{eq:pose_annotations}
\end{equation}
The number of people $P$ may vary between CSI observations.

To accommodate a variable number of people within a fixed
decoder structure, WiSPER uses $Q$ pose prediction slots.
Each slot produces a complete 3D pose candidate and a
presence logit, from which a presence score is obtained.
The person index $p\in\{1,\ldots,P\}$ identifies an annotated
person, whereas the slot index $q\in\{1,\ldots,Q\}$ identifies
a model prediction. These indices have no predefined
correspondence, and not every prediction slot is expected
to represent a person.

Let $F_{\Theta}$ denote the WiSPER inference model
with trainable parameters $\Theta$. The model maps a CSI
observation to a set of $Q$ pose candidates and their
presence logits:
\begin{equation}
    F_{\Theta}(\mathbf{X})
    =
    \left\{
        \left(\widehat{\mathbf{Y}}_q,b_q\right)
    \right\}_{q=1}^{Q},
    \label{eq:task}
\end{equation}
where $\widehat{\mathbf{Y}}_q\in\mathbb{R}^{K\times3}$ is
the predicted pose in slot $q$, and $b_q\in\mathbb{R}$ is
its presence logit.
The corresponding presence score is
$s_q=\operatorname{sigmoid}(b_q)$, with a higher score
indicating greater predicted confidence that the slot
represents a person.
The slots correspond to pose queries in the Transformer
decoder, allowing the model to generate pose candidates
in parallel without receiving the number of people
as an input.

\subsection{Our Proposed WiSPER}
Figure~\ref{fig:overview} summarizes WiSPER.
We first describe PAMEL and then the transfer of its
context encoder to ReFT.

\noindent
\textbf{Stage 1: PAMEL.}
For supervised predictive pretraining, we tokenize CSI
by projecting the $C$ channels at each time--link pair
into a $D$-dimensional embedding and adding temporal
and link positional encodings. Following
WiFi-JEPA~\cite{kim2026wifijepa}, we mask entire links
across time. With visible and masked indices denoted
by $V$ and $M$, the context encoder produces
$\mathbf{Z}_{V}=f_{\theta}(\mathbf{X}_{V})$.
The latent predictor estimates
$\widehat{\mathbf{Z}}_{M}=p_{\phi}(\mathbf{Z}_{V},M)$
using visible embeddings and positioned mask tokens.
The full-observation target encoder supplies
$\mathbf{Z}^{\star}_{M}
=\operatorname{sg}[
\operatorname{LN}(f_{\bar{\theta}}(\mathbf{X}))_{M}]$,
where $\operatorname{LN}$ denotes layer normalization
and $\operatorname{sg}$ stops gradients.
The latent objective is
\begin{equation}
    \mathcal{L}_{\mathrm{JEPA}}
    =
    \frac{1}{|M|D}
    \sum_{i\in M}\sum_{d=1}^{D}
    \ell_{\mathrm{SL1}}
    (\widehat{Z}_{i,d}-Z^{\star}_{i,d}),
    \label{eq:pamel_latent_loss}
\end{equation}
where $\ell_{\mathrm{SL1}}$ is Smooth-$L_1$ loss with
unit transition. Target parameters follow
$\bar{\theta}\leftarrow m\bar{\theta}+(1-m)\theta$.

An auxiliary head predicts
$\{(\widehat{\mathbf{Y}}^{\mathrm{aux}}_r,
b^{\mathrm{aux}}_r)\}_{r=1}^{A}
=a_{\omega}(\mathbf{Z}_{V})$,
where $A$ is the number of pretraining pose slots.
A one-to-one assignment matches annotations to predictions
using presence and coordinate costs.
The supervised loss $\mathcal{L}_{\mathrm{pose}}$ combines
focal loss over all slots with $L_1$ coordinate loss
over matched poses. Both objectives therefore act on
the same visible-context representation:
\begin{equation}
    \mathcal{L}_{\mathrm{PAMEL}}
    =
    \mathcal{L}_{\mathrm{JEPA}}
    +\lambda\mathcal{L}_{\mathrm{pose}}.
    \label{eq:pamel_objective}
\end{equation}
We periodically estimate
\begin{equation}
\lambda^{\star}=\operatorname{clip}
(\|\nabla_{\theta}\mathcal{L}_{\mathrm{JEPA}}\|_2/
\|\nabla_{\theta}\mathcal{L}_{\mathrm{pose}}\|_2,
\lambda_{\min},\lambda_{\max})
\end{equation}
for positive gradient norms.
The first valid estimate initializes $\lambda$,
and subsequent estimates update it through an
exponential moving average.
Only the context encoder is retained for ReFT.

\noindent
\textbf{Stage 2: ReFT}
The transferred encoder and hierarchical pose
Transformer~\cite{shi2022end} process full CSI to produce
$\{(\mathbf{Y}^{\mathrm{c}}_q,\mathbf{h}_q,b_q)\}_{q=1}^{Q}
=d_{\eta}(f_{\theta}(\mathbf{X}))$,
where $\mathbf{Y}^{\mathrm{c}}_q\in\mathbb{R}^{K\times3}$
is a coarse pose, $\mathbf{h}_q$ contains per-joint
features, and $b_q$ is its presence logit.
The residual network is conditioned on
$\mathbf{c}_q=(\operatorname{sg}[\mathbf{Y}^{\mathrm{c}}_q],
\operatorname{sg}[\mathbf{h}_q])$.
The coordinates provide the reference geometry,
while the features supply CSI-dependent information.

A one-to-one training assignment matches person $p$
to slot $q_p$, defining the normalized correction
$\mathbf{z}_{1,p}
=(\mathbf{Y}_p-\operatorname{sg}[\mathbf{Y}^{\mathrm{c}}_{q_p}])/\sigma$,
where $\sigma>0$ is a fixed scale.
We sample $\mathbf{z}_{0,p}$ with independent standard
normal entries and $t\sim U[0,1)$, constructing
$\mathbf{z}_{t,p}=(1-t)\mathbf{z}_{0,p}+t\mathbf{z}_{1,p}$.
Following the linear-path formulation of flow
matching~\cite{lipman2023flow}, we parameterize refinement
through an endpoint network $g_{\psi}$ with attention
across joint tokens:
\begin{equation}
    \mathcal{L}_{\mathrm{FM}}
    =
    \mathbb{E}\!\left[
        \frac{
        \|g_{\psi}(\mathbf{z}_{t,p},t,\mathbf{c}_{q_p})
        -\mathbf{z}_{1,p}\|_F^2
        }{3K}
    \right].
    \label{eq:reft_endpoint_loss}
\end{equation}
The expectation covers matched pairs, noise, and flow
time. The endpoint prediction defines
$v_{\psi}(\mathbf{z},t,\mathbf{c})
=(g_{\psi}(\mathbf{z},t,\mathbf{c})-\mathbf{z})
/\max(1-t,\varepsilon)$.
Where the clamp is inactive, endpoint squared error
equals $(1-t)^2$ times velocity squared error against
$\mathbf{z}_{1,p}-\mathbf{z}_{0,p}$.
Thus, the objective uses endpoint regression with
an implicit time weighting.

The complete ReFT objective is
\begin{equation}
    \mathcal{L}_{\mathrm{ReFT}}
    =
    \mathcal{L}_{\mathrm{set}}
    +\mathcal{L}_{\mathrm{FM}},
    \label{eq:reft_objective}
\end{equation}
where $\mathcal{L}_{\mathrm{set}}$ supervises presence
and coordinates across the Transformer's proposal,
pose-decoder, and joint-refinement outputs.
It updates the encoder and pose Transformer;
the detached residual objective updates only $g_{\psi}$.

At inference, we integrate $H$ independent Gaussian
initializations using $S$ Euler steps of size $1/S$,
with fixed conditioning and the endpoint network
evaluated at each step. The final residual states
produce
\begin{equation}
    \widehat{\mathbf{Y}}_q
    =
    \mathbf{Y}^{\mathrm{c}}_q
    +\frac{\sigma}{H}
    \sum_{n=1}^{H}\mathbf{z}^{(n)}_{S,q}.
    \label{eq:reft_refined_pose}
\end{equation}
Presence logits remain unchanged.

\begin{table}[t]
\centering
\caption{Single-person comparison on all 2,586 PiW3D test
frames}
\label{tab:single_person_baselines}
\resizebox{\columnwidth}{!}{%
\begin{tabular}{lrrr}
\toprule
Method 
& MPJPE (mm) $\downarrow$
& PA-MPJPE (mm) $\downarrow$
& PCK@0.2 (\%) $\uparrow$
\\
\midrule
GraphPose-Fi \cite{chen2026graph}
& 622.38 & 120.56 & 6.64 
\\
DT-Pose \cite{chen2026dt}
& 91.86 & 61.14 & 72.07 
\\
WiFi-JEPA \cite{kim2026wifijepa}
& 81.20 & 57.08 & 74.26 
\\
\textbf{Ours WiSPER}
& \textbf{49.42} & \textbf{35.68} & \textbf{89.07} 
\\
\bottomrule
\end{tabular}%
}
\end{table}
\begin{table}[t]
\centering
\caption{PiW3D comparison across person counts. All entries report MPJPE in mm
(lower is better). The one-, two-, and three-person subsets contain
2,586, 3,184, and 2,054 frames, respectively.}
\label{tab:multi_person_baselines}
\resizebox{.8\columnwidth}{!}{%
\begin{tabular}{lrrrr}
\toprule
Method & 1 person & 2 people & 3 people  & Average \\
\midrule
WiFi-JEPA \cite{kim2026wifijepa} & 81.20 & 110.10 & 131.93  & 106.28 \\
WiSPER   & 49.42 & 63.76  & 81.65   & 63.72  \\
\bottomrule
\end{tabular}}
\end{table}

\begin{figure*}[t]
 \centering
 \includegraphics[width=.9\textwidth]{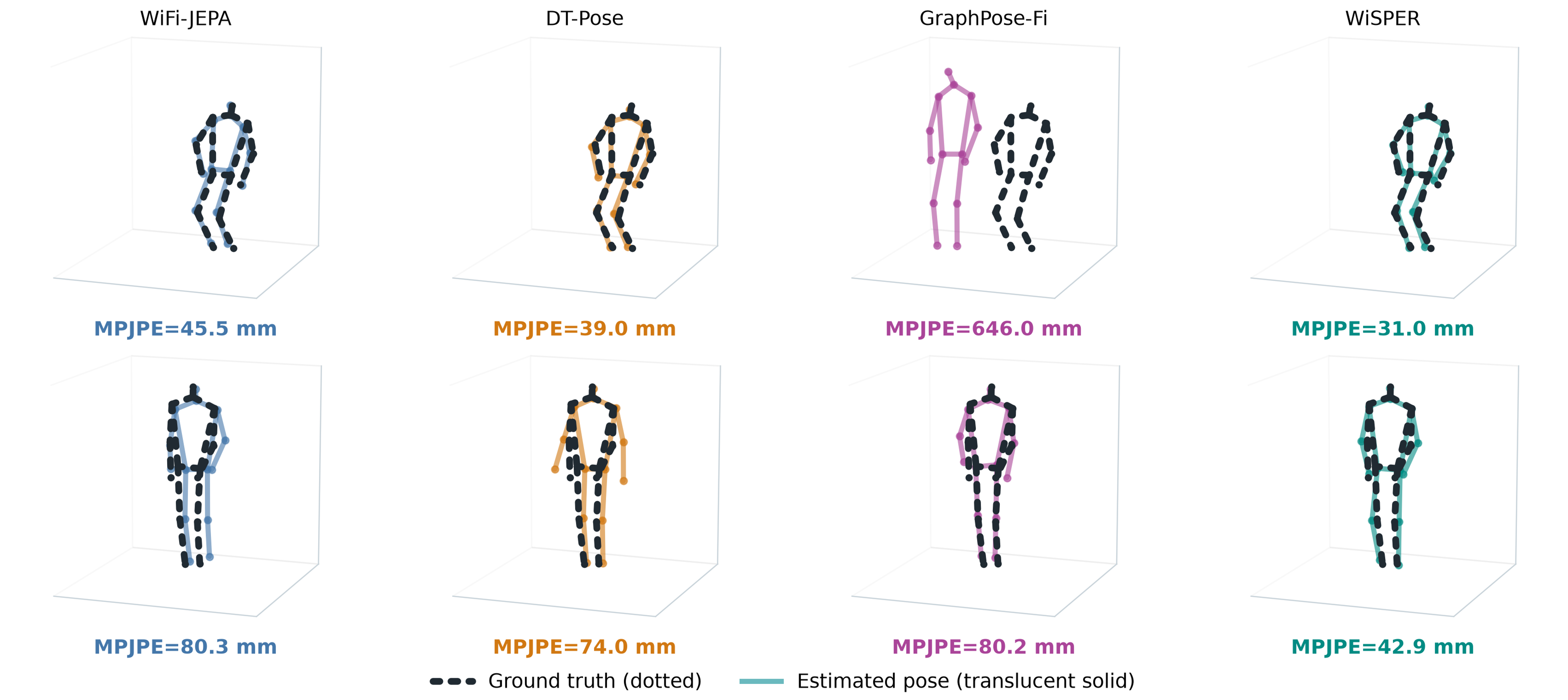}
 \caption{Single-person pose comparisons on two PiW3D test
observations. Each row shows the same observation across
methods. Ground truth is shown with thick dark dotted lines,
and estimated poses with thinner translucent solid lines.
The value below each panel is the MPJPE for that observation. }
 \label{fig:viz}
\end{figure*}
\begin{table*}[t]
\centering
\caption{Component ablations on PiW3D. All errors are in mm (lower is better).
ReFT (Flow) Off bypasses residual flow refinement at inference;
the pose Transformer remains active. 
Pose-only pretraining also changes the auxiliary loss
weighting scheme.}
\label{tab:component_ablation}
\resizebox{\textwidth}{!}{%
\begin{tabular}{lllrrrrr}
\toprule
& & & \multicolumn{4}{c}{MPJPE} & PA-MPJPE \\
\cmidrule(lr){4-7}
\cmidrule(lr){8-8}
Ablated Component  & Pretraining & ReFT (Flow)
& 1 person & 2 people & 3 people & All & All \\
\midrule
Auxiliary pose supervision; flow refinement
& CSI-only JEPA & Off
& 84.02 & 113.99 & 133.91 & 109.31 & 69.02 \\
Auxiliary pose supervision
& CSI-only JEPA & On
& 73.68 & 94.56 & 112.00 & 92.24 & 56.49 \\
\midrule
Masked latent prediction; flow refinement
& Pose-only (fixed weight) & Off
& 57.67 & 76.15 & 99.87 & 76.27 & 53.06 \\
Masked latent prediction
& Pose-only (fixed weight) & On
& \textbf{49.36} & 65.20 & 84.96 & 65.15 & 43.01 \\
\midrule
Flow refinement
& PAMEL & Off
& 57.20 & 74.24 & 94.55 & 73.94 & 51.53 \\
None (full WiSPER)
& PAMEL & On
& 49.42 & \textbf{63.76} & \textbf{81.65} & \textbf{63.72} & \textbf{42.27} \\
\bottomrule
\end{tabular}%
}
\vspace{-0.3cm}
\end{table*}
\begin{table}[t]
\centering
\caption{Sampling configurations evaluated using the same
WiSPER checkpoint.
Each row reports one full-test evaluation.}
\label{tab:sampling_ablation}
\begin{tabular}{rrr}
\toprule
Trajectories $H$ & Euler steps $S$ & MPJPE (mm) \\
\midrule
1  & 1  & 64.71 \\
1  & 5  & 67.32 \\
8  & 5  & 63.72 \\
32 & 5  & 63.33 \\
8  & 10 & 63.98 \\
\bottomrule
\end{tabular}
\end{table}
\section{Experiments and Evaluations} \label{sec:exp}

\noindent
\textbf{Dataset}
We evaluate on Person-in-WiFi 3D
(PiW3D)~\cite{person3dyan}, which provides synchronized
CSI and 14-joint pose annotations for seven volunteers
performing eight actions across three indoor environments.
The released split contains 89,946 training frames and
7,824 test frames, including 2,586 single-person,
3,184 two-person, and 2,054 three-person observations.
Both WiSPER stages use the paired training data across
all person counts. Training and test frames originate
from different temporal portions of shared recordings;
this evaluation therefore does not establish generalization
to unseen participants or environments.

\noindent
\textbf{Implementation.}
CSI inputs have dimensions $C\times T\times L=60\times20\times9$,
comprising amplitude and sanitized phase.
The encoder has 12 layers, dimension 512, and eight
attention heads. PAMEL masks five of nine links and
uses an eight-layer latent predictor and a two-layer
auxiliary pose head with $A=3$ slots.
The pose Transformer uses $Q=100$ queries, dimension 256,
eight heads, three pose-decoder layers, and two
joint-refinement layers. The residual network has
three layers, dimension 256, and four heads.
Both training stages use AdamW for 100 epochs with
batch size 64 and seed 42. Pretraining uses a learning
rate of $3\times10^{-4}$ and weight decay 0.05.
Fine-tuning uses a base learning rate of $2\times10^{-5}$
and weight decay $10^{-4}$, with encoder and residual-network
learning-rate multipliers of 0.1 and 5, respectively.
The PAMEL loss weight is updated every 50 steps,
with bounds $[10^{-4},1]$ and EMA coefficient 0.9.
ReFT uses $\sigma=0.05$~m, $\varepsilon=10^{-3}$,
$H=8$ trajectories, and $S=5$ Euler steps.
Results are from individual training runs;
variation across training seeds is not evaluated.

\noindent
\textbf{Baselines}
We compare WiSPER with three baselines: WiFi-JEPA~\cite{kim2026wifijepa},
DT-Pose~\cite{chen2026dt}, and
GraphPose-Fi~\cite{chen2026graph}.
WiFi-JEPA provides the multi-person comparison;
while DT-Pose and GraphPose-Fi pose estimators
are trained on single-person observations.
Note that GraphPose-Fi is adapted from the official implementation
to nine CSI links and a 14-joint graph, and
uses amplitude-only inputs.

\noindent
\textbf{Evaluation Metrics}
We report mean per-joint position error (MPJPE) and
Procrustes-aligned MPJPE (PA-MPJPE) in millimetres.
MPJPE measures absolute-coordinate error; PA-MPJPE
removes translation, rotation, and uniform scale.
Errors are averaged over joints and people within
each frame, then across frames.
We report normalized PCK@0.2, counting joints whose
error is within 20\% of the annotated left--right
elbow distance. This definition differs from PCK
with a fixed millimetre threshold.

\noindent
\textbf{Results: Single-person Pose Estimation}
Table~\ref{tab:single_person_baselines} reports results
on 2,586 single-person test frames.
WiSPER achieves an MPJPE
of 49.42~mm and PA-MPJPE of 35.68~mm, compared with
81.20~mm and 57.08~mm for WiFi-JEPA.
PCK@0.2 increases from 74.26\% to 89.07\%.
DT-Pose and the local GraphPose-Fi adaptation obtain
direct-output MPJPE values of 91.86~mm and 622.38~mm,
respectively. GraphPose-Fi's substantially lower
PA-MPJPE of 120.56~mm indicates that global alignment
contributes to its error; this result characterizes
our PiW3D adaptation rather than its published performance.
Figure~\ref{fig:viz} illustrates the selected predictions
on shared test observations.

\noindent
\textbf{Results: Multi-person Pose Estimation}
Table~\ref{tab:multi_person_baselines} compares WiSPER with
WiFi-JEPA for which we have completed multi-person
evaluations under the same local protocol.
The DT-Pose and GraphPose-Fi implementations were trained and evaluated
for single-person estimation.
WiSPER reduces MPJPE from 110.10~mm to 63.76~mm for
two-person observations and from 131.93~mm to 81.65~mm
for three-person observations. Both methods exhibit
increasing error as the number of people grows, consistent
with the greater difficulty of recovering multiple
skeletons from shared CSI observations. Nevertheless,
WiSPER maintains lower error across all person counts,
extending its single-person improvement to multi-person
estimation.

\noindent
\textbf{Ablation study}
Table~\ref{tab:component_ablation} compares three
pretraining configurations with residual refinement
enabled or bypassed at inference.
The pose-supervised configurations obtain lower errors
than CSI-only JEPA. With refinement, PAMEL achieves
63.72~mm overall MPJPE, compared with 65.15~mm for
pose-only pretraining, a modest 2.2\% reduction.
PAMEL performs better on two- and three-person
observations, while pose-only pretraining is slightly
better on single-person observations
(49.36 versus 49.42~mm).
Because these configurations also differ in loss
weighting, and results are from individual training
runs, this comparison does not isolate the contribution
of masked latent prediction.
Applying the trained residual refiner reduces overall
MPJPE by 15.6\%, 14.6\%, and 13.8\% for CSI-only JEPA,
pose-only pretraining, and PAMEL, respectively.
These consistent improvements support applying the
refiner within the evaluated models, but do not
establish superiority over direct residual regression.

\noindent
\textbf{Flow sampling.}
Table~\ref{tab:sampling_ablation} varies the trajectory
count $H$ and Euler-step count $S$ at the same checkpoint.
A single trajectory and step achieves 64.71~mm MPJPE,
compared with 63.72~mm for the main configuration
($H=8$, $S=5$).
Increasing $H$ to 32 yields 63.33~mm, whereas increasing
$S$ to 10 with $H=8$ yields 63.98~mm.
More integration steps therefore do not consistently
improve accuracy in these evaluations.

\noindent\textbf{Limitations.}
Our current experimental results
are based on individual training runs, and the
reported matched-pose errors do not assess
person-count accuracy. Evaluation
uses temporal portions of shared recordings,
leaving generalization to unseen participants
and environments, as well as inference latency,
for future investigation.

\section{Conclusions} \label{sec:conclusions}
WiSPER reduces overall MPJPE by 40.0\% relative to
WiFi-JEPA on PiW3D dataset,
with reductions of 42.1\% and 38.1\% for two- and
three-person observations, respectively. 
Within WiSPER, residual flow refinement further reduces
overall MPJPE by 13.8\% compared with bypassing refinement
at inference, demonstrating the benefit of correcting
the remaining coordinate errors.
Future work
will focus on generalization evaluation across environments
and unseen participants, and reducing paired supervision
through independently pretrained CSI representations
and human-pose priors.

\bibliographystyle{IEEEbib}
\bibliography{refs}

\end{document}